\documentclass[letterpaper]{article}
\usepackage{aaai20}
\usepackage{times}
\usepackage{helvet}
\usepackage{courier}
\usepackage[hyphens]{url}
\usepackage{graphicx}
\usepackage{graphicx}
\usepackage{amsmath}
\usepackage{amssymb}
\usepackage{multirow}
\DeclareMathOperator*{\argmax}{arg\,max}
\DeclareMathOperator*{\argmin}{arg\,min}
\title{Ladder Loss for Coherent Visual-Semantic Embedding}
\author{Mo Zhou\textsuperscript{1},~
Zhenxing Niu\textsuperscript{2},~
Le Wang\textsuperscript{3},~
Zhanning Gao\textsuperscript{2,3},~
Qilin Zhang\textsuperscript{4},~
Gang Hua\textsuperscript{5}\\
\textsuperscript{1}Xidian University,~
\textsuperscript{2}Alibaba Group,~
\textsuperscript{3}Xi'an Jiaotong University,~
\textsuperscript{4}HERE Technologies,~
\textsuperscript{5}Wormpex AI Research\\
{\tt\small
\{cdluminate,zhenxingniu,zhanninggao,samqzhang,ganghua\}@gmail.com,
lewang@xjtu.edu.cn
}
}

\begin{document}

\maketitle
\begin{abstract}

	For visual-semantic embedding, the existing methods normally treat the
	relevance between queries and candidates in a bipolar way -- relevant or
	irrelevant, and all ``irrelevant'' candidates are uniformly pushed away
	from the query by an equal margin in the embedding space, regardless of
	their various proximity to the query.  This practice disregards relatively
	discriminative information and could lead to suboptimal ranking in the
	retrieval results and poorer user experience, especially in the long-tail
	query scenario where a matching candidate may not necessarily exist.
	In this paper, we introduce
	a continuous variable to model the relevance degree between queries and
	multiple candidates, and propose to learn a coherent embedding space,
	where candidates with higher relevance degrees are mapped closer to the query
	than those with lower relevance degrees. In particular, the new ladder loss is
	proposed by extending the triplet loss inequality to a more general
	inequality chain, which implements variable push-away margins according to
	respective relevance degrees.  In addition, a proper Coherent Score
	metric is proposed to better measure the ranking results including those
	``irrelevant'' candidates. Extensive experiments on multiple datasets validate
	the efficacy of our proposed method, which achieves significant improvement
	over existing state-of-the-art methods.

\end{abstract}


%
\section{Introduction}
\begin{figure}[!t]
\centering
\includegraphics[width=1.0\columnwidth]{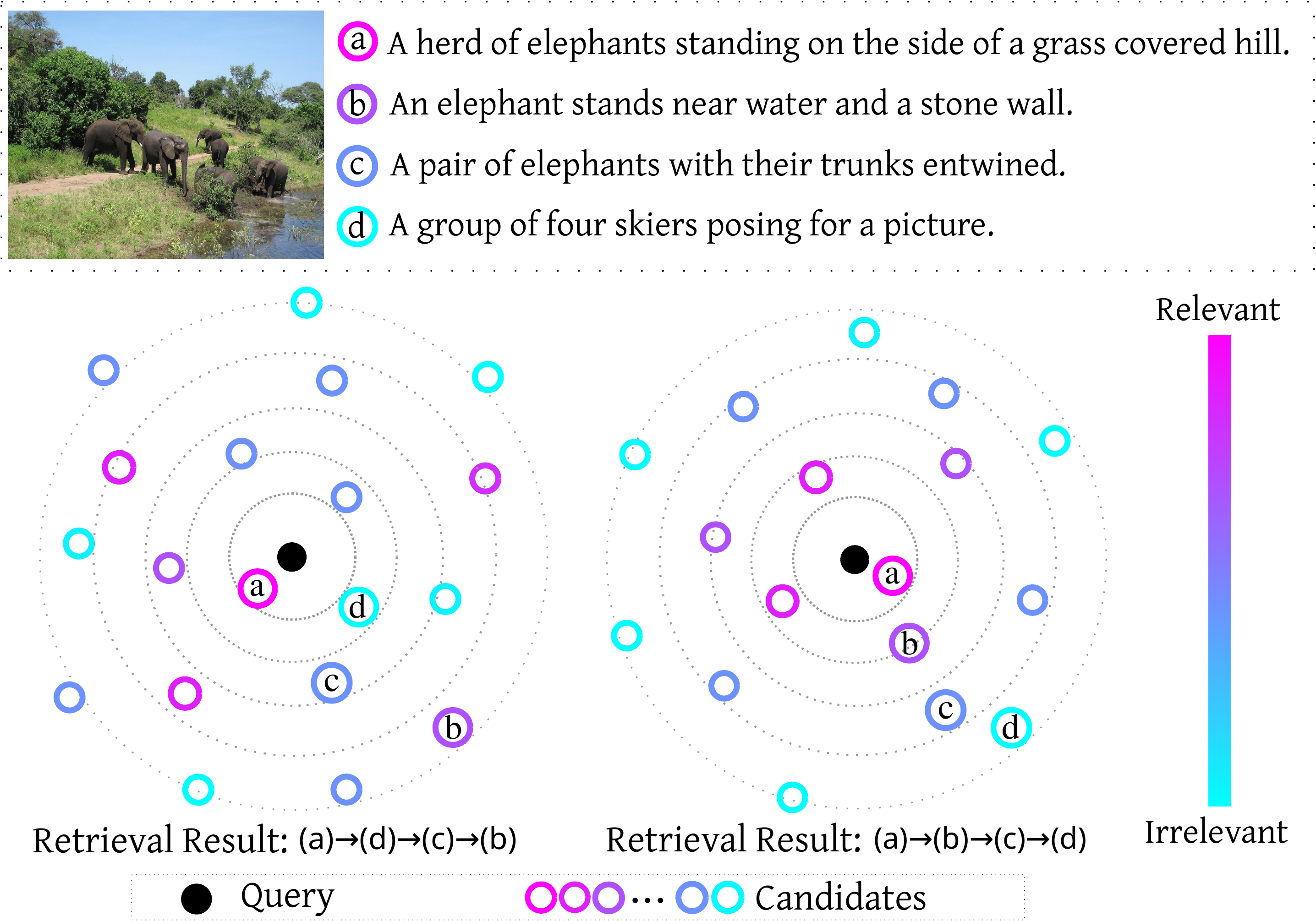}
\caption{Comparison between the incoherent (left) and coherent (right)
    visual-semantic embedding space. Existing methods (left) pull the
    totally-relevant sentence (a) close to the query image, while pushing away all
    other sentences (b, c, and d) equally. Therefore, the
    relative proximity of (b, c, and d) are not necessarily consistent with
    their relevance degrees to the query (solid black dot). On contrary,
    our approach (right) explicitly preserves the proper relevance order in the
    retrieval results.} \label{fig_first}
\end{figure}
Visual-semantic embedding aims to map images and their descriptive sentences
into a common space, so that we can retrieve sentences given query images or
vice versa, which is namely cross-modal retrieval~\cite{ji2017cross}.
Recently, the advances in deep learning have made significant progress on
visual-semantic embedding~\cite{Kiros1,Karpathy1,Karpathy2,VSEPP}.  Generally,
images are represented by the Convolutional Neural Networks (CNN), and
sentences are represented by the Recurrent Neural Networks (RNN). A triplet
ranking loss is subsequently optimized to make the corresponding
representations as close as possible in the embedding
space~\cite{schroff2015facenet,sohn2016improved}.

For visual-semantic embedding, previous
methods~\cite{hadsell2006dimensionality,schroff2015facenet} tend to treat the
relevance between queries and candidates in a bipolar way: for a query image,
only the corresponding ground-truth sentence is regarded as \textbf{relevant},
and other sentences are \emph{equally} regarded as \textbf{irrelevant}.
Therefore, with the triplet ranking loss, only the relevant sentence is pulled
close to the query image, while all the irrelevant sentences are pushed away
\emph{equally}, \emph{i.e.}, be pushed from the query by an equal margin.
However, among those so-called \textbf{irrelevant} sentences, some are more 
relevant to the query than others, thus should be treated accordingly.

Similarly, it is arguably a disadvantage in recent retrieval evaluation
metrics which disregard the ordering/ranking of retrieved ``irrelevant'' results.
For example, the most popular Recall@K (\emph{i.e.}, R@K)~\cite{Kiros1,Karpathy1,VSEPP} 
is purely based on the ranking position of the ground-truth candidates (denoted as
\emph{totally-relevant} candidates in this paper); while \emph{neglecting} the ranking order
of all other candidates. 
However, the user experience of a practical cross-modal retrieval system could
be heavily impacted by the ranking order of all top-$N$ candidates, including the
``irrelevant'' ones, as it is often challenging to retrieve enough totally-relevant
candidates in the top-$N$ results (known as the long-tail query challenge~\cite{downey2007heads}).
Given a query from the user, when a exact matching candidate does not exist in
the database, a model trained with only bipolar supervision information will
likely fail to retrieve those somewhat relevant candidates, and produce a badly ordered ranking result.
As demonstrated in Fig.~\ref{fig_first}, given a query
image (solid black dot), the ground-truth sentence (a) is the totally-relevant one,
which does occupy the top of the retrieved list. Besides that,
the sentence (b) is notably more relevant than (c) or (d), so 
ideally the (b) should be ranked before the (c), and
the (d) should be ranked at the bottom.

Therefore, it is beneficial to formulate the semantic \textbf{relevance degree}
as a continuous variable rather than a binary variable (\emph{i.e.}, relevant
or irrelevant). And the relevance degree should be incorporated into
embedding space learning, so that the candidates with
higher relevance degrees will be closer to the query than those with lower
degrees.

In this paper, we first propose to measure the relevance degree between images
and sentences, based on which we design the \textbf{ladder loss} to learn a
\emph{coherent} embedding space. The ``coherent'' means that the similarities
between queries and candidates are conformal with their relevance degrees.
Specifically, the similarity between the query image $i_q$ and its
totally-relevant sentence $t_q$ in the conventional triplet loss~\cite{VSEPP}
is encouraged to be greater than the similarity between the $i_q$ and other
sentences $t_p$. Likewise, with the ladder loss formulation, we consider the
relevance degrees of all sentences, and extend the inequality $s(i_q,
t_q)>s(i_q, t_p)$ to an inequality chain, \emph{i.e.}, $s(i_q,
t_q)>s(i_q,t_{p_1})>s(i_q,t_{p_2})>\dots>s(i_q,t_{p_L})$, where $t_{p_l}$ is more
relevant to $i_q$ than $t_{p_{l+1}}$, and $s(\cdot,\cdot)$ denotes cosine similarity.
Using the inequality chain, we design the ladder loss so that the sentences with lower relevance
degrees will be pushed away by a larger margin than the ones with higher
relevance degrees. As a result, it leads to learn a coherent embedding space, and both
the totally-relevant as well as the somewhat-relevant sentences can be properly ranked.

In order to better evaluate the quality of retrieval results, we
propose a new \textbf{Coherent Score (CS)} metric, which is designed
to measure the alignment between the real ranking order and the expected
ranking order. The expected ranking order is decided according to the relevance
degrees, so that the CS can properly reflect user experience for cross-modal
retrieval results. In brief, our contributions are:
\begin{enumerate}
\item We propose to formulate the relevance degree as a continuous rather than a
	binary variable, which leads to learn a coherent embedding space, where
	both the totally-relevant and the somewhat-relevant candidates can be
	retrieved and ranked in a proper order.
\item To learn a coherent embedding space, a ladder loss is proposed by
	extending the inequality in the triplet loss to an inequality chain, so
	that candidates with different degrees will be treated differently.
\item A new metric, Coherent Score (CS), is proposed to evaluate the ranking
	results, which can better reflect user experience in a cross-modal
	retrieval system.
\end{enumerate}
\section{Related Work}
\textbf{Visual-semantic Embedding}, as a kind of multi-modal joint embedding,
enables a wide range of tasks in image and language understanding, such as
image-caption retrieval~\cite{Karpathy2,Kiros1,VSEPP}, image captioning, and
visual question-answering~\cite{Malinowski_2015_ICCV}. Generally, the methods
of visual-semantic embedding could be divided into two categories. The first
category is based on Canonical Correlation Analysis (CCA)
\cite{hardoon2004canonical,gong2014multi,gong2014improving,klein2014fisher}
which finds linear projections that maximize the correlation between projected
vectors from the two modalities. Extensions of CCA to a deep learning framework
have also been proposed \cite{andrew2013deep,yan2015deep}.

The second category involves metric learning-based embedding space
learning ~\cite{Frome,DeepSP,VSEPP}.  DeViSE~\cite{Frome,Socher2} learns linear
transformations of visual and textual features to the common space. After that,
Deep Structure-Preserving (DeepSP)~\cite{DeepSP} is proposed for image-text
embedding, which combines cross-view ranking constraints with within-view
neighborhood structure preservation. In \cite{Niu2017}, Niu {\em et al.} propose to learn a
hierarchical multimodal embedding space where not only full sentences and
images but also phrases and image regions are mapped into the space. Recently,
Fartash {\em et al.} \cite{VSEPP} incorporate hard negatives in the ranking loss
function, which yields significant gains in retrieval performance.
Compared to CCA-based methods, metric learning-based methods scale better
to large dataset with stochastic optimization in training.

\textbf{Metric learning}, has many other applications such as face
recognition~\cite{schroff2015facenet} and fine-grained
recognition~\cite{oh2016deep,wu2017sampling,yuan2017hard}.
The loss function design in metric learning could be a subtle problem.
For example, the contrastive loss~\cite{hadsell2006dimensionality} 
pulls all positives close, while all negatives are separated by a fixed distance.
However, it could be severely restrictive to enforce such fixed distance
for all negatives.
This motivated the triplet loss~\cite{schroff2015facenet}, which only requires
negatives to be farther away than any positives on a per-example basis,
\emph{i.e.}, a less restrictive relative distance constraint.
After that, many variants of triplet loss are proposed.
For example, PDDM~\cite{huang2016local} and Histogram
Loss~\cite{ustinova2016learning} use quadruplets. Beyond that, the n-pair
loss~\cite{sohn2016improved} and Lifted Structure~\cite{oh2016deep} define
constraints on all images in a batch. However, all the aforementioned methods
formulate the relevance as a binary variable.
Thus, our ladder loss could be used to boost those methods.
\section{Our Approach}
Given a set of image-sentence pairs $\mathcal{D}=\{(i_n,t_n)_{n=1}^N\}$,
the visual-semantic embedding aims to map both images $\{(i_n)_{n=1}^N\}$ and
sentences $\{(t_n)_{n=1}^N\}$ into a common space. In previous methods, for
each image $i_q$, only the corresponding sentence $t_q$ is regarded as
relevant, and the others $\{t_p, (p\in \mathcal{N}^{-q})\}$ are all regarded as
irrelevant, where $\mathcal{N}^{-q}=\{n|1\leq n \leq N, \text{and } n\neq q\}$.
Thus, only the inequality $s(i_q,t_q)>s(i_q,t_p), (p\in \mathcal{N}^{-q})$ is
enforced in previous methods.

In contrast, our approach will measure the semantic relevance degree between
$i_q$ and each sentence in $\{t_p, (p\in \mathcal{N}^{-q})\}$. Intuitively, the
corresponding sentence $t_q$ should have the highest relevance degree, while
the others would have different degrees. Thus, in our coherent embedding space,
the similarity of an image-sentence pair with higher relevance degree is desired
to be greater than the similarity for a pair with lower degree.

To this end, we first define a continuous variable to measure the semantic
relevance degree between images and sentences (in Sec.~\ref{SRD}). Subsequently,
to learn a coherent embedding space, we design a novel ladder loss to push
different candidates away by distinct margins according to their relevance
degree (in Sec.~\ref{LL}). At last, we propose the Coherent Score metric to
properly measure whether the ranking order is aligned with their relevance degrees
(in Sec.~\ref{CS}).

Our approach only relies on customized loss function and it has no restrictions
on the image/sentence representation, so it is flexible to be incorporated into
any neural network architecture.
\subsection{Relevance Degree} \label{SRD}
In our approach, we need to measure the semantic relevance degree for
image-sentence pairs. The ideal ground-truth for image-sentence pair is human annotation,
but in fact it is infeasible to annotate such a multi-modal pairwise
relevance dataset due to the combinatorial explosion in the number of possible pairs.
On the other hand, the single-modal relevance measurement (\emph{i.e.}, between sentences) is often much
easier than the cross-modal one (\emph{i.e.}, between sentences and images).
For example, recently many newly proposed Natural Language Processing (NLP) models~\cite{devlin2018bert,ELMo,MTDNN}
achieved very impressive results~\cite{glue} on various NLP tasks.
Specifically, on the sentence similarity task the BERT~\cite{devlin2018bert}
has nearly reached human performance. Compared to single-modal metric learning
in image modality, the natural language similarity measure is more mature.
Hence we cast the image-sentence relevance problem as a sentence-sentence relevance problem.

Intuitively, for an image $i_q$, the relevance degree of
its corresponding sentence $t_q$ is supposed to be the highest, and it is regarded
as a reference when measuring the relevance degrees between $i_q$ and other sentences.
In other words, measuring the relevance degree between the image $i_q$ and the sentence $t_p,~(p\in
\mathcal{N})$ is cast as measuring the relevance degree (i.e. similarity) between the two sentences $t_q$ and $t_p,
~(p\in \mathcal{N})$.

To this end, 
we employ the Bidirectional Encoder Representations Transformers (BERT)~\cite{devlin2018bert}.
Specifically, the BERT model we used is fine-tuned on the Semantic Textual Similarity
Benchmark (STS-B) dataset\cite{2017STS,devlin2018bert}.
The Pearson correlation coefficient of our fine-tuned BERT on STS-B validation set is $0.88$,
which indicates good alignment between predictions and human perception.
In short, the relevance degree between an image $i_q$ and a sentence $t_p$ is
calculated as the similarity score between $t_q$ and $t_p$ with our fine-tuned BERT model:
\begin{equation}
  R(i_q,t_p) = R(t_q,t_p)= \text{BERT}(t_q, t_p).\label{eq:bertrd}
\end{equation}
\subsection{Ladder Loss Function} \label{LL}
In this section, the conventional triplet loss is briefly overviewed, followed
by our proposed ladder loss.
\subsubsection{Triplet Loss}
Let $v_q$ be the visual representation of a query image $i_q$, and $h_p$
indicates the representation of the sentence $t_p$. In the triplet loss formulation, for
query image $i_q$, only its corresponding sentence $t_q$ is regarded as
the positive (\emph{i.e.}, relevant) sample; while all other sentences $\{t_p, (p\in
\mathcal{N}^{-q})\}$ are deemed negative (\emph{i.e.}, irrelevant).  Therefore,
in the embedding space the similarity between $v_q$ and $h_q$ is encouraged to be
greater than the similarity between $v_q$ and $h_p, (p\in \mathcal{N}^{-q})$ by a
margin $\alpha$, 
\begin{equation}
  s(v_q,h_q) - s(v_q,h_p) > \alpha, (p\in \mathcal{N}^{-q}) , \label{ieq_t}
\end{equation}
which can be transformed as the triplet loss function,
\begin{equation}
 L_{tri}(q) = \sum_{p\in \mathcal{N}^{-q} } [\alpha- s(v_q,h_q) + s(v_q,h_p)]_+ , \label{loss_t}
\end{equation}
where $[x]_+$ indicates $\max\{0, x\}$.
Considering the reflexive property of the query and candidate, the full triplet loss is
\begin{equation}
\begin{aligned}
\mathcal{L}_{tri}(q) = & \sum_{p\in N^{-q} } [\alpha- s(v_q,h_q) + s(v_q,h_p)]_+ \\
+ & \sum_{p\in N^{-q} } [\alpha- s(h_q,v_q) + s(h_q,v_p)]_+ . \label{tri_full}
\end{aligned}
\end{equation}
\subsubsection{Ladder Loss}
\begin{figure*}[t!]
\centering
\includegraphics[width=0.95\linewidth]{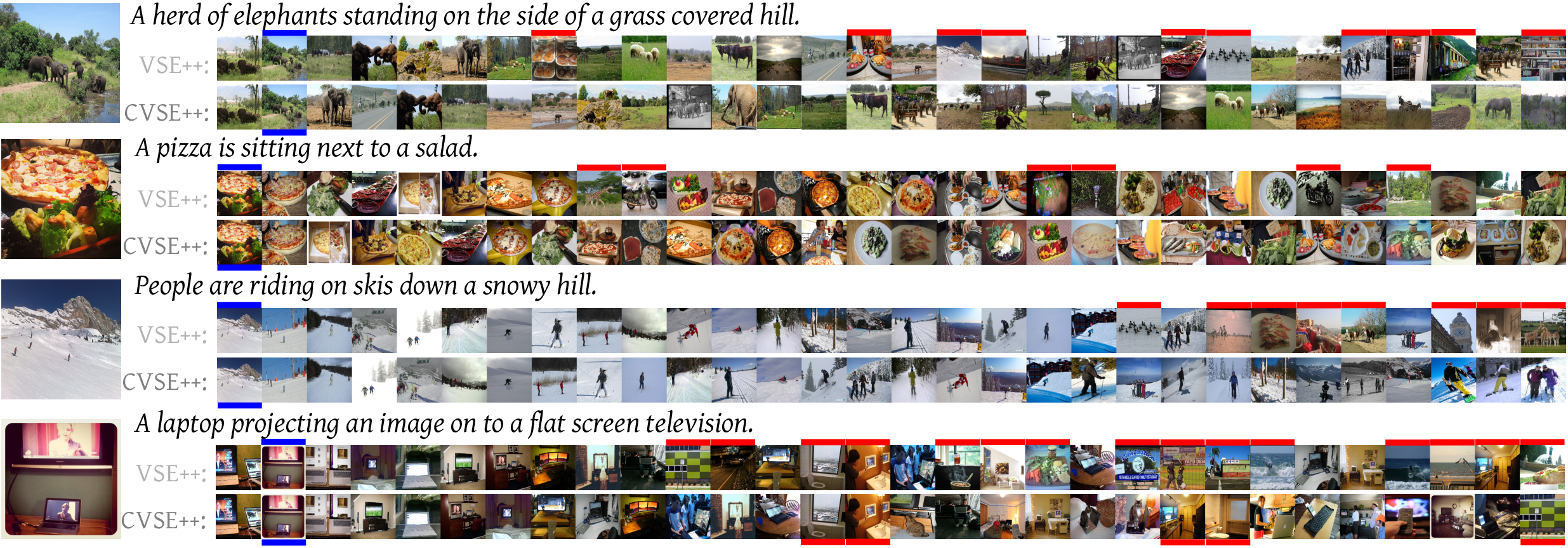}
\caption{Comparison of the sentence-to-image top-$30$ retrieval results between VSE++
	(baseline, $1$st row) and CVSE++ (Ours, $2$nd row).
	For each query sentence, the ground-truth image is shown on the left,
	the totally-relevant and totally-irrelevant retrieval results are
	marked by blue and red overlines/underlines, respectively.
	Despite that both methods retrieve the totally-relevant images at
	identical ranking positions, the baseline VSE++ method includes
    more totally-irrelevant images in the top-$30$ results; while our
    proposed CVSE++ method mitigates such problem.
	} \label{fig_vis2}
\end{figure*}
We first calculate the relevance degrees
between image $i_q$ and each sentence $t_p, (p\in \mathcal{N}^{-q})$. After
that, these relevance degree values are divided into $L$ levels with
thresholds $\theta_l, (l=1,2,\dots,L-1)$. As a result, the sentence index set
$\mathcal{N}^{-q}$ is divided into $L$ subsets
$\mathcal{N}^{-q}_1,\mathcal{N}^{-q}_2,\dots,\mathcal{N}^{-q}_L$, and 
sentences in $\mathcal{N}^{-q}_{l}$ are more relevant to the query than the
sentences in $\mathcal{N}^{-q}_{l+1}$.

To learn a coherent embedding space, the more relevant sentences should be
pulled closer to the query than the less relevant ones. To this end, we
extend the single inequality Eq.~\eqref{ieq_t} to an inequality chain,
\begin{equation}
\begin{aligned}
s(v_q,h_q) - s(v_q,h_i) & > \alpha_1, (i\in \mathcal{N}^{-q}_1), \\
s(v_q,h_i) - s(v_q,h_j) & > \alpha_2, (i\in \mathcal{N}^{-q}_1, j\in \mathcal{N}^{-q}_2), \\
s(v_q,h_j) - s(v_q,h_k) & > \alpha_3, (j\in \mathcal{N}^{-q}_2, k\in \mathcal{N}^{-q}_3), \\
& \cdots,
\end{aligned}
\end{equation}
where $\alpha_1,\dots,\alpha_L$ are the margins between different non-overlapping sentence subsets.

In this way, the sentences with distinct relevance degrees are pushed away
by distinct margins. For examples, for
sentences in $\mathcal{N}^{-q}_1$, they are pushed away by margin $\alpha_1$,
and for sentences in $\mathcal{N}^{-q}_2$, they are pushed away by margin
$\alpha_1+\alpha_2$.
Based on such inequality chain, we could define the ladder loss function. For
simplicity, we just show the ladder loss with three-subset-partition
(\emph{i.e.}, $L=3$) as an example, 
\begin{eqnarray}
& L_{lad}(q) = \beta_1 L_{lad}^1(q) + \beta_2 L_{lad}^2(q) + \beta_3 L_{lad}^3(q),  \label{loss_tradeoff} \\
& L_{lad}^1(q) = \sum_{i\in \mathcal{N}^{-q}_{1:L}} [\alpha_1- s(v_q,h_q) +
s(v_q,h_i)]_+ \nonumber, \\
& L_{lad}^2(q) = \sum_{i\in \mathcal{N}^{-q}_1, j\in \mathcal{N}^{-q}_{2:L}} [\alpha_2- s(v_q,h_i) + s(v_q,h_j)]_+ , \label{loss_lad_2} \\
& L_{lad}^3(q) = \sum_{j\in \mathcal{N}^{-q}_2, k\in \mathcal{N}^{-q}_{3:L}} [\alpha_3- s(v_q,h_j) + s(v_q,h_k)]_+ \nonumber ,
\end{eqnarray}
where $\beta_1$, $\beta_2$ and $\beta_3$ are the weights between
$L_{lad}^1(q)$, $L_{lad}^2(q)$ and $L_{lad}^3(q)$, respectively. $\mathcal{N}^{-q}_{l:L}$
indicates the union from $\mathcal{N}^{-q}_l$ to $\mathcal{N}^{-q}_L$.

As can be expected, the $L_{lad}^1(q)$ term alone is identical to the original
triplet loss, {\em i.e.}, the ladder loss degenerates to the triplet loss if
$\beta_2=\beta_3=0$.
Note that the dual problem of sentence as a query and images as candidates also exists.
Similar to obtaining the full triplet loss Eq.~\eqref{tri_full},
we can easily write the full ladder loss $\mathcal{L}_{lad}(q)$, which is omitted here.
\subsubsection{Ladder Loss with Hard Contrastive Sampling}
For visual-semantic embedding, the hard negative sampling
strategy~\cite{simo2015discriminative,wu2017sampling} has been
validated for inducing significant performance improvements,
where selected hard samples (instead of all samples) are utilized
for the loss computation.
Inspired by~\cite{wu2017sampling,VSEPP}, we develop a similar strategy of
selecting hard contrastive pairs for the ladder loss computation,
which is termed \textbf{hard contrastive sampling (HC)}.

Taking the $L_{lad}^2(q)$ in Eq.~\eqref{loss_lad_2} as an example, instead of
conducting the sum over the sets $i\in \mathcal{N}^{-q}_1$ and $j\in
\mathcal{N}^{-q}_{2:L}$, we sample one or several pairs $(h_i,h_j)$ from
$i\in \mathcal{N}^{-q}_1$ and $j\in \mathcal{N}^{-q}_{2:L}$.
Our proposed HC sampling strategy involves choosing
the $h_j$ closest to the query in $\mathcal{N}^{-q}_{2:L}$, and the
$h_i$ furthest to the query in $\mathcal{N}^{-q}_1$ for the loss computation.
Thus, the ladder loss
part $L_{lad}^2(q)$ with hard contrastive sampling can be written as,
\begin{equation}
\begin{aligned}
L_{lad-HC}^2(q) &= [\alpha_1- s(v_q,h_{i^*}) + s(v_q,h_{j^*})]_+ ,\\
j^* &= \argmax_{j\in \mathcal{N}^{-q}_{2:L}}{s(v_q,h_j)} ,\\
i^* &= \argmin_{i\in \mathcal{N}^{-q}_1}{s(v_q,h_i)} ,
\end{aligned}
\end{equation}
where $(i^*,j^*)$ is the index of the hardest contrastive pair $(h_{i^*},h_{j^*})$.
According to our empirical observation, this HC strategy not only reduces the
complexity of loss computation, but also improves the overall performance.
\begin{table*}[!t]
\centering
\resizebox{\textwidth}{!}{%
\begin{tabular}{|c|c|c|c|c|c|c|c|c|c|c|c|c|}
\hline
\multicolumn{13}{|c|}{MS-COCO (1000 Test Samples)}\tabularnewline
\hline
\multirow{2}{*}{Model} & \multicolumn{6}{c|}{Image$\rightarrow$Sentence} & \multicolumn{6}{c|}{Sentence$\rightarrow$Image}\tabularnewline
\cline{2-13} \cline{3-13} \cline{4-13} \cline{5-13} \cline{6-13} \cline{7-13} \cline{8-13} \cline{9-13} \cline{10-13} \cline{11-13} \cline{12-13} \cline{13-13}
 & CS@100 & CS@1000 & Mean R & R@1 & R@5 & R@10 & CS@100 & CS@1000 & Mean R & R@1 & R@5 & R@10\tabularnewline
\hline
Random & 0.018 & 0.009 & 929.9 & 0.0 & 0.3 & 0.5 & 0.044 & 0.005 & 501.0 & 0.1 & 0.5 & 0.9\tabularnewline
\hline
VSE++ (VGG19) & 0.235 & 0.057 & 5.7 & 56.7 & 83.9 & 92.0 & 0.237 & 0.057 & 9.1 & 42.6 & 76.5 & 86.8\tabularnewline
\hline
CVSE++ (VGG19) & 0.256 & 0.347 & 4.1 & 56.8 & 83.6 & 92.2 & 0.257 & 0.223 & 7.3 & 43.2 & 77.5 & 88.1\tabularnewline
\hline
VSE++ (VGG19,FT) & 0.253 & 0.047 & 2.9 & 62.5 & 88.2 & 95.2 & 0.246 & 0.042 & 6.5 & 49.9 & 82.8 & 91.2\tabularnewline
\hline
CVSE++ (VGG19,FT) & 0.256 & 0.419 & 2.8 & 63.2 & 89.9 & 95.0 & 0.251 & 0.287 & 5.3 & 50.5 & 83.6 & 92.8\tabularnewline
\hline
VSE++ (Res152) & 0.238 & 0.079 & 2.8 & 63.2 & 88.9 & 95.5 & 0.236 & 0.080 & 7.3 & 47.4 & 80.3 & 89.9\tabularnewline
\hline
CVSE++ (Res152) & 0.265 & 0.358 & 2.8 & 66.7 & 90.2 & 94.0 & 0.256 & 0.236 & 6.1 & 48.4 & 81.0 & 90.0\tabularnewline
\hline
VSE++ (Res152,FT) & 0.241 & 0.071 & 2.4 & 68.0 & 91.9 & 97.4 & 0.239 & 0.068 & 6.3 & 53.5 & 85.1 & 92.5\tabularnewline
\hline
CVSE++ (Res152,FT) & 0.265 & 0.446 & 2.4 & 69.1 & 92.2 & 96.1 & 0.255 & 0.275 & 4.7 & 55.6 & 86.7 & 93.8\tabularnewline
\hline
\hline
\multicolumn{13}{|c|}{MS-COCO (5000 Test Samples)}\tabularnewline
\hline
\multirow{2}{*}{Model} & \multicolumn{6}{c|}{Image$\rightarrow$Sentence} & \multicolumn{6}{c|}{Sentence$\rightarrow$Image}\tabularnewline
\cline{2-13} \cline{3-13} \cline{4-13} \cline{5-13} \cline{6-13} \cline{7-13} \cline{8-13} \cline{9-13} \cline{10-13} \cline{11-13} \cline{12-13} \cline{13-13}
 & CS@500 & CS@5000 & Mean R & R@1 & R@5 & R@10 & CS@500 & CS@5000 & Mean R & R@1 & R@5 & R@10\tabularnewline
\hline
VSE++ (Res152) & 0.227 & 0.078 & 10.6 & 36.3 & 66.8 & 78.7 & 0.224 & 0.084 & 30.9 & 25.6 & 54.0 & 66.9\tabularnewline
\hline
CVSE++ (Res152) & 0.253 & 0.354 & 9.7 & 39.3 & 69.1 & 80.3 & 0.246 & 0.239 & 25.2 & 25.8 & 54.0 & 67.3\tabularnewline
\hline
VSE++ (Res152,FT) & 0.231 & 0.073 & 7.7 & 40.2 & 72.5 & 83.3 & 0.228 & 0.073 & 25.1 & 30.7 & 60.7 & 73.3 \tabularnewline
\hline
CVSE++ (Res152,FT) & 0.255 & 0.439 & 7.4 & 43.2 & 73.5 & 84.1 & 0.242 & 0.280 & 18.6 & 32.4 & 62.2 & 74.6\tabularnewline
\hline
\end{tabular}
}
\caption{Comparison between VSE++ and CVSE++ in terms of CS@K and R@K on MS-COCO.} \label{tab_coco}
\end{table*}
\subsection{Coherent Score} \label{CS}
In previous methods, the most popular metric for visual-semantic embedding is
R@K, which only accounts for the ranking position of the ground-truth candidates
(\emph{i.e.}, the totally-relevant candidates) while neglects others. Therefore,
we propose a novel metric Coherent Score (CS) to properly measure the
ranking order of all top-$N$ candidates (including the ground-truth and other
candidates).

The CS@K is defined to measure the alignment between the real
ranking list $r_1,r_2,\dots,r_K$ and its expected ranking list
$e_1,e_2,\dots,e_K$, where thee expected ranking list is decided according to their
relevance degrees.
We adopt Kendall's rank correlation
coefficient $\tau,~(\tau\in[-1,1])$~\cite{kendall} as the criterion.
Specifically, any pair of $(r_i,e_i)$ and $(r_j,e_j)$ where $i<j$ is defined to be
concordant if both $r_i>r_j$ and $e_i>e_j$, or if both $r_i<r_j$ and
$e_i<e_j$. Conversely, it is defined to be discordant if the ranks for
both elements mismatch. The Kendall's rank correlation $\tau$ depends
on the number of concordant pairs and discordant pairs.
When $\tau=1$, the alignment is perfect, \emph{i.e.} the two ranking lists are identical.
Thus,
a high CS@K score indicates the good quality and good user experience of the learnt embedding space
and retrieval result in terms of coherence, and a model that achieves high CS@K
score is expected to perform better in long-tail query challenges~\cite{downey2007heads}
where a perfect match to the query does not necessarily exist in the database.
\section{Experiments} \label{EXP}
\begin{table*}[!t]
\centering
\resizebox{0.95\textwidth}{!}{%
\begin{tabular}{|c|c|c|c|c|c|c|c|c|c|c|c|c|}
\hline
\multirow{2}{*}{Model} & \multicolumn{6}{c|}{Image$\rightarrow$Sentence} & \multicolumn{6}{c|}{Sentence$\rightarrow$Image}\tabularnewline
\cline{2-13} \cline{3-13} \cline{4-13} \cline{5-13} \cline{6-13} \cline{7-13} \cline{8-13} \cline{9-13} \cline{10-13} \cline{11-13} \cline{12-13} \cline{13-13}
 & CS@100 & CS@1000 & Mean R & R@1 & R@5 & R@10 & CS@100 & CS@1000 & Mean R & R@1 & R@5 & R@10\tabularnewline
\hline
\hline
Random & 0.02 & -0.005 & 988.3 & 0.0 & 0.3 & 0.4 & -0.033 & -0.003 & 503.0 & 0.2 & 0.6 & 1.1\tabularnewline
\hline
VSE++ (VGG19) & 0.116 & 0.139 & 18.2 & 40.7 & 68.4 & 78.0 & 0.115 & 0.124 & 26.9 & 28.7 & 58.6 & 69.8 \tabularnewline
\hline
CVSE++ (VGG19) & 0.129 & 0.255 & 16.4 & 42.8 & 69.2 & 78.9 & 0.127 & 0.144 & 26.4 & 29.0 & 59.2 & 71.1\tabularnewline
\hline
VSE++ (VGG19,FT) & 0.128 & 0.130 & 14.7 & 44.6 & 73.3 & 82.0 & 0.125 & 0.110 & 22.8 & 31.9 & 63.0 & 74.5\tabularnewline
\hline
CVSE++ (VGG19,FT) & 0.133 & 0.260 & 13.0 & 44.8 & 73.1 & 82.3 & 0.131 & 0.160 & 20.8 & 33.8 & 63.9 & 75.1\tabularnewline
\hline
VSE++ (Res152) & 0.126 & 0.127 & 10.2 & 49.3 & 78.9 & 86.4 & 0.115 & 0.112 & 20.0 & 35.9 & 65.9 & 75.6\tabularnewline
\hline
CVSE++ (Res152) & 0.133 & 0.247 & 9.3 & 50.2 & 78.8 & 87.3 & 0.120 & 0.147 & 20.0 & 37.1 & 66.9 & 76.4\tabularnewline
\hline
VSE++ (Res152,FT) & 0.130 & 0.122 & 7.8 & 54.1 & 81.0 & 88.7 & 0.122 & 0.114 & 16.2 & 39.8 & 70.0 & 79.0\tabularnewline
\hline
CVSE++ (Res152,FT) & 0.141  & 0.273 & 7.4 & 56.6 & 82.5 & 90.2 & 0.126 & 0.172 & 15.7 & 42.4 & 71.6 & 80.8\tabularnewline
\hline
\end{tabular}
}
\caption{Comparison between VSE++ and CVSE++ in terms of CS@K and R@K on Flickr30K.} \label{tab_f30k}
\end{table*}
Following related works, Flickr30K~\cite{Flickr30k} and MS-COCO~\cite{coco,coco2} datasets are used in
our experiments. The two datasets contain $31,000$ and $123,000$ images, respectively,
and each image within them is annotated with $5$ sentences using AMT. For
Flickr30K, we use $1,000$ images for validation, $1,000$ for testing and the
rest for training, which is consistent with \cite{VSEPP}. For MS-COCO, we also
follow \cite{VSEPP} and use $5,000$ images for both validation and testing.
Meanwhile, the rest $30,504$ images in original
validation set are used for training ($113,287$ training images in total) in our experiments following~\cite{VSEPP}.
Our experimental settings follow that in VSE++~\cite{VSEPP}, which is the
state-of-the-art for visual-semantic embedding. Note, in terms of image-sentence
cross modal retrieval, SCAN~\cite{SCAN} achieves better performance, but it
does not learn a joint embedding space for full sentences and full images, and
suffers from combinatorial explosion in the number of sample pairs to be evaluated.

VGG-19~\cite{VGG} or ResNet-152~\cite{He2015resnet}-based image representation
is used for our experiments (both pre-trained on ImageNet).
Following common practice, we extract $4096$ or $2048$-dimensional feature vectors
directly from the penultimate fully connected layer from these networks.
We also adopt random cropping in data augmentation, where all images are
first resized to $256\times 256$ and randomly cropped $10$ times at $224\times 224$ resolution.
For the sentence representation, we use a Gated Recurrent Unit (GRU),
similar to the one used in \cite{VSEPP}. The dimension of the GRU
and the joint embedding space is set at $D=1024$. The dimension of the word embeddings
used as input to the GRU is set to $300$.

Additionally, Adam solver is used for optimization, with the learning rate set
at \verb|2e-4| for $15$ epochs, and then decayed to \verb|2e-5| for another 15
epochs. We use a mini-batch of size $128$ in all experiments in this paper.
Our algorithm is implemented in PyTorch~\cite{paszke2017automatic}.
\subsection{Relevance Degree} \label{exp_srd}
The BERT inference is highly computational expensive ({\em e.g.}, a single
NVIDIA Titan Xp GPU could compute similarity score for only approximately $65$ sentence pairs per second).
Therefore, it is computational infeasible to directly use Eq.~\eqref{eq:bertrd} in practice
due to combinatorial explosion of the number of sentence pairs.

In this paper, we mitigate the problem by introducing a coarse-to-fine
mechanism. For each sentence pair we first employ conventional
CBoW~\cite{glue} method to coarsely measure their relevance degree. If the
value is larger than a predefined threshold, Eq.~\eqref{eq:bertrd} is used to
refine their relevance degree calculation. The CBoW method first
calculates each sentence's representation by averaging the GloVe~\cite{glove}
word vectors for all tokens, and then computes the cosine similarity
between their representations of each sentence pair.
With this mechanism, the false-positive ``relevant'' pairs found by the CBoW
method would be suppressed by BERT, while those important real relevant pairs
would be assigned with more accurate relevance degrees.
Thus, the speed of CBoW and the accuracy of BERT are combined properly.
We empirically fix the predefined threshold at $0.8$ for our experiments,
as the mechanism achieves $0.79$ in person correlation on STS-B.
\subsection{Results on MS-COCO} \label{exp_coco}
We compare VSE++ (re-implemented) and our Coherent Visual-Semantic Embedding (CVSE++) on the MS-COCO dataset, where VSE++ only
focuses on the ranking position of the totally-relevant candidates while our
approach cares about the ranking order of all Top-$N$ candidates.  
The method of VSE++~\cite{VSEPP} is our baseline since it is the
state-of-the-art approach for learning visual-semantic embedding. 
For fair comparison, we use both Recall@K (denoted as ``R@K'') and CS@K as metrics for evaluation,
and also fine-tune (denoted by ``FT'') the CNNs following the baseline.
In our approach, the hard contrastive sampling strategy is used. Experiments
without the hard negative or hard contrastive sampling strategy are omitted
because they perform much worse in terms of R@K, as reported in \cite{VSEPP}.

In our approach, we need to determine the ladder number $L$ in the
loss function, which depends on how many top-ranked candidates (the value of $N$)
we care about (\emph{i.e.}, termed the scope-of-interest in this paper).
With a small scope-of-interest, \emph{e.g.}, top-$100$, only a few ladders
are required, \emph{e.g.}, $L=2$; but with a larger scope-of-interest,
\emph{e.g.}, top-$200$, we will need more ladders, \emph{e.g.}, $L=3$,
so that the low-level ladder, \emph{e.g.}, $L_{lad}^2(q)$ in Eq.~\eqref{loss_tradeoff},
is responsible for optimizing the ranking order of the very top candidates,
\emph{e.g.}, top-$1$ $\sim$ top-$100$; while the high-level ladder,
\emph{e.g.}, $L_{lad}^3(q)$ in Eq.~\eqref{loss_tradeoff}, is responsible for optimizing
the ranking order of subsequent candidates, \emph{e.g.}, top-$100$ $\sim$ top-$200$.

A detailed discussion regarding the scope-of-interest and the choice of
ladder number $L$ will be provided in the next section.
Practically, we limit our illustrated results to $L=2$ both for computational savings
and for the limited scope-of-interest from most human users.
With ladder number $L$ fixed at $2$, parameters can be empirically
determined by exploiting the validation set, {\em e.g.}, the threshold $\theta_1$
for splitting $\mathcal{N}^{-q}_1$ and $\mathcal{N}^{-q}_2$ is fixed at $0.63$, and the
margins $\alpha_{1}=0.2$, $\alpha _2=0.01$, the loss weights $\beta_1=1$, $\beta_2=0.25$.

With our proposed CS@K metric, significantly larger $K$ values are chosen than those
({\em e.g.}, $1, 5, 10$) in the classical R@K metric. For instance, we report the CS@100
and CS@1000 with 1000 test samples.
Such choices of $K$ allow more insights into both the local and global order-preserving effects in
embedding space. In addition, the conventional R@K metrics are also included to
measure the ranking performance of the totally-relevant candidates.

\begin{table*}[ht!]
\centering
\resizebox{0.9\textwidth}{!}{%
\begin{tabular}{|c|c|c|c|c|c|c|c|c|c|c|c|c|}
\hline
\multirow{2}{*}{$\beta_2$} & \multicolumn{6}{c|}{Image$\rightarrow$Sentence} & \multicolumn{6}{c|}{Sentence$\rightarrow$Image}\tabularnewline
\cline{2-13} \cline{3-13} \cline{4-13} \cline{5-13} \cline{6-13} \cline{7-13} \cline{8-13} \cline{9-13} \cline{10-13} \cline{11-13} \cline{12-13} \cline{13-13}
 & CS@100 & CS@1000 & Mean R & R@1 & R@5 & R@10 & CS@100 & CS@1000 & Mean R & R@1 & R@5 & R@10\tabularnewline
\hline
\hline
0.0 & 0.238 & 0.079 & 2.8 & 63.2 & 88.9 & 95.5 & 0.236 & 0.08 & 7.3 & 47.4 & 80.3 & 89.9\tabularnewline
\hline
0.25 & 0.265 & 0.358 & 2.8 & 66.7 & 90.2 & 94.0 & 0.256 & 0.236 & 6.1 & 48.4 & 81.0 & 90.0\tabularnewline
\hline
1.0 & 0.266 & 0.417 & 3.9 & 64.0 & 88.2 & 93.1 & 0.259 & 0.264 & 6.2 & 47.4 & 79.0 & 88.9 \tabularnewline
\hline
\end{tabular}
}
\caption{Performance of the proposed CVSE++(Res152) with respect to
	the parameter $\beta_2$ (On MS-COCO dataset).} \label{tab_beta}
\end{table*}
\begin{table*}[ht!]
\centering
\resizebox{1.0\textwidth}{!}{%
\begin{tabular}{|c|c|c|c|c|c|c|c|c|c|c|c|c|c|c|}
\hline
\multirow{2}{*}{L} & \multicolumn{7}{c|}{Image$\rightarrow$Sentence} & \multicolumn{7}{c|}{Sentence$\rightarrow$Image}\tabularnewline
\cline{2-15} \cline{3-15} \cline{4-15} \cline{5-15} \cline{6-15} \cline{7-15} \cline{8-15} \cline{9-15} \cline{10-15} \cline{11-15} \cline{12-15} \cline{13-15} \cline{14-15} \cline{15-15}
 & CS@100 & CS@200 & CS@1000 & Mean R & R@1 & R@5 & R@10 & CS@100 & CS@200 & CS@1000 & Mean R & R@1 & R@5 & R@10\tabularnewline
\hline
\hline
1 & 0.238 & 0.188 & 0.079 & 2.8 & 63.2 & 88.9 & 95.5 & 0.236 & 0.189 & 0.08 & 7.3 & 47.4 & 80.3 & 89.9\tabularnewline
\hline
2 & 0.265 & 0.252 & 0.358 & 2.8 & 66.7 & 90.2 & 94.0 & 0.256 & 0.253 & 0.236 & 6.1 & 48.4 & 81.0 & 90.0\tabularnewline
\hline
3 & 0.267 & 0.274  & 0.405 & 3.2 & 65.7 & 89.3 & 94.1 & 0.261 & 0.258 & 0.244  & 6.3 & 48.4 & 80.3 & 89.4\tabularnewline
\hline
\end{tabular}
}
\caption{Performance of the proposed CVSE++(Res152) with
	respect to the ladder number $L$. (On MS-COCO dataset)} \label{tab_numlad}
\end{table*}
The experimental results on the MS-COCO dataset are presented in Tab.~\ref{tab_coco},
where the proposed CVSE++ approaches evidently outperform their corresponding VSE++ counterparts
in terms of CS@K, \emph{e.g.},
from VSE++(Res152): $0.238$ to CVSE++(Res152): $0.265$ in terms of CS@100 for image$\rightarrow$sentence retrieval with 1000 MS-COCO test samples.
Moreover, the performance improvements are more significant with the larger
scope-of-interest at CS@1000, \emph{e.g.}, where ``CVSE++ (Res152,FT)'' achieves over $5$-fold
increase over ``VSE++ (Res152,FT)'' (from $0.071$ to $0.446$) in
image$\rightarrow$sentence retrieval. The result indicates that with our proposed ladder
loss a coherent embedding space could be effectively learnt, which could
produce significantly better ranking results especially in the global scope.

Simultaneously, a less expected phenomenon can be observed from Tab.~\ref{tab_coco}:
our proposed CVSE++ variants achieve roughly comparable or marginally better
performance than their VSE++ counterparts in terms of R@K,
\emph{e.g.}, from VSE++(Res152): $63.2$ to CVSE++(Res152): $66.7$ in terms of R@1 for image$\rightarrow$sentence retrieval with 1000 MS-COCO test samples.
The overall improvement in R@K is insignificant because it completely neglects the ranking
position of those non-ground-truth samples, and CVSE++ is not designed for improving the ranking for ground-truth.
Based on these results, we speculate that the ladder loss
appears to be beneficial (or at least not harmful) to the inference of totally-relevant candidates.
Nevertheless,
there are still hyper-parameters ($\beta_1, \beta_2, \cdots, \beta_L$) controlling the balance between
the totally-relevant and somewhat-relevant candidates, which will be further analyzed in the next section.

To provide some visual comparison between VSE++ and CVSE++,
several sentences are randomly sampled 
from the validation set as queries, and their corresponding retrievals are illustrated 
in Fig.~\ref{fig_vis2} (sentence$\rightarrow$image). 
Evidently, our CSVE++ could
put more somewhat-relevant candidates and reduce the number of totally-irrelevant candidates 
on the top-$N$ retrieval list and enhance user experience.
\subsection{Results on Flickr30K}
Our approach is also evaluated on the Flikr30K dataset and compared with
the baseline VSE++ variants, as shown in Tab.~\ref{tab_f30k}.
The hyper-parameter settings are identical to that in Tab.~\ref{tab_coco}
with MS-COO (1000 Test Samples).
As expected, these experimental results demonstrate similar performance
improvements both in terms of CS@K and R@K by our proposed CVSE++ variants.
\section{Parameter Sensitivity Analysis} 
In this section, parameter sensitivity analysis is carried out on two groups
of hyper-parameters, {\em i.e.}, the balancing parameter
$\beta_1, \beta_2, \cdots, \beta_L$ in Eq.~\eqref{loss_tradeoff} and the ladder number $L$.
\subsection{Balancing Totally Relevant and Others} \label{exp_beta}
In Eq.~\eqref{loss_tradeoff}, the weights between the ranking position
optimization of totally-relevant candidates and other candidates in the
ladder loss are controlled by the hyper-parameters $\beta_1, \beta_2, \cdots, \beta_L$.
With $\beta_2=\cdots=\beta_L=0$,  the ladder loss degenerates to the triplet loss,
and all emphasis is put on the totally-relevant ones. Conversely,
relatively larger $\beta_2, \cdots, \beta_L$ values put more emphasis on
the somewhat-relevant candidates.

With other parameters fixed ($L$ fixed at $2$, $\beta_1$ fixed at $1$),
parameter sensitivity analysis is carried out on $\beta_2$ only.
From Tab.~\ref{tab_beta}, we can see that CS@K metrics improve with larger $\beta_2$,
but R@K metrics degrade when $\beta_2$ is close to $1.0$.
Based on the three $\beta_2$ settings in Tab.~\ref{tab_beta},
we speculate that CS@K and R@K metrics would not necessarily
peak simultaneously at the same $\beta_2$ value.
We also observe that with excessively large $\beta_2$ values,
the R@K metrics drop dramatically. Generally, the ranking orders of the
totally-relevant candidates often catch user's attention and
they should be optimized with high priority.
Therefore, we select $\beta_2=0.25$ in all our other experiments
to strike a balance because of R@K and CS@K performance.
%
\subsection{The Scope-of-interest for Ladder Loss} \label{sec_discuss}
Our approach focuses on improving the ranking order of all top-$N$ retrieved results
(instead of just the totally-relevant ones).  Thus, there is an important parameter,
\emph{i.e.}, the scope-of-interest $N$ or the size of the desired retrieval list.
If the retrieval system user only cares about a few top-ranked results
(\emph{e.g.}, top-$100$), two ladders (\emph{e.g.}, $L=2$) are practically sufficient;
If a larger scope-of-interest (\emph{e.g.}, top-$200$) is required,
more ladders are probably needed in the ladder loss. For example, with $L=3$,
the low-level ladder $L_{lad}^2(q)$ is responsible for the optimization
of the ranking order of very top candidates, \emph{e.g.}, from top-$1$ $\sim$ top-$100$;
while the high-level ladder $L_{lad}^3(q)$ is responsible for the optimization of
the ranking order of subsequent candidates, \emph{e.g.}, from top-$100$ $\sim$ top-$200$.
Inevitably, larger ladder number results in higher computational complexity.
Therefore, a compromise between the scope-of-interest and the computational complexity
needs to be reached.

For the sensitivity analysis of ladder number $L = 1, 2, 3$, we evaluate our
CVSE++ (Res152) approach by comparing top-$100$, top-$200$ and top-$1000$ results,
which are measured by CS@100, CS@200 and CS@1000, respectively.
Other parameters $\theta_2$, $\alpha_3$, $\beta_3$ are empirically
fixed at $0.56$, $0.01$, $0.125$, respectively.
The experimental results are summarized in Tab.~\ref{tab_numlad}.
With small scope-of-interest $N=100$, we find that two ladder $L=2$ is effective
to optimize the CS@100 metric, a third ladder only incurs marginal improvements.
However, with larger scope-of-interest, {\em e.g.}, top-$200$,
the CS@200 can be further improved by adding one more ladder, {\em i.e.}, $L=3$.

Apart from that, a notable side effect with too many ladders (\emph{e.g.} $5$) can be observed,
the R@K performance drops evidently.
We speculate that with more ladders, the ladder loss is likely to
be dominated by high-level ladder terms and leads to some difficulties
in optimization of the low-level ladder term. This result indicates
that the choice of $L$ should be proportional to the scope-of-interest,
\emph{i.e.}, more ladders for larger scope-of-interest and vice versa.
\section{Conclusion}
In this paper, relevance between queries and candidates are formulated as a
continuous variable instead of a binary one, and a new ladder loss is proposed
to push different candidates away by distinct margins. As a result, we could
learn a coherent visual-semantic space where both the totally-relevant and
the somewhat-relevant candidates can be retrieved and ranked in a proper
order.

In particular, our ladder loss improves the ranking quality of all top-$N$
results without degrading the ranking positions of the ground-truth candidates. Besides,
the scope-of-interest is flexible by adjusting the number of ladders. Extensive
experiments on multiple datasets validate the efficacy of our proposed method,
and our approach achieves the state-of-the-art performance in terms of both
CS@K and R@K. For future work, we plan to extend the 
ladder loss-based embedding to other metric learning applications.

\subsection{Acknowledgements}

This work was supported partly by National Key R\&D Program of China Grant
2018AAA0101400, NSFC Grants 61629301, 61773312, 61976171, and 61672402. China
Postdoctoral Science Foundation Grant 2019M653642, and Young Elite Scientists
Sponsorship Program by CAST Grant 2018QNRC001.

%
{\small
\bibliographystyle{aaai}
\bibliography{egbib}
}

\end{document}